\documentclass{article}

\usepackage{PRIMEarxiv}

\usepackage[table]{xcolor}  
\usepackage[utf8]{inputenc}
\usepackage{geometry}
\usepackage{graphicx}
\usepackage{caption}
\usepackage{float}
\usepackage{booktabs}
\usepackage{appendix}
\usepackage{hyperref}
\usepackage{rotating}
\usepackage{multirow}
\usepackage{svg}
\usepackage{pdflscape}
\usepackage[colorinlistoftodos]{todonotes}
\usepackage{subcaption}
\definecolor{lightyellow}{RGB}{255, 255, 204} 
\usepackage{tabularx}
\usepackage{enumitem}
\usepackage[utf8]{inputenc} 
\usepackage[T1]{fontenc}    
\usepackage{hyperref}       
\usepackage{url}            
\usepackage{booktabs}       
\usepackage{amsfonts}       
\usepackage{nicefrac}       
\usepackage{microtype}      
\usepackage{lipsum}
\usepackage{fancyhdr}       
\usepackage{graphicx}       
\graphicspath{{images/}}     

\title{Experimenting AI Technologies for Disinformation Combat: the IDMO Project
}

\author{
  Lorenzo Canale, Alberto Messina \\
  Rai - Radiotelevisione Italiana  \\
  Centro Ricerche, Innovazione Tecnologica e Sperimentazione (CRITS) \\
  Via Giovanni Carlo Cavalli 6, 10129, Torino\\
  \texttt{\{lorenzo.canale, alberto.messina\}@rai.it} \\
}

\begin{document}
\maketitle

\begin{abstract}
The Italian Digital Media Observatory (IDMO) project, part of a European initiative, focuses on countering disinformation and fake news. This report outlines contributions from Rai-CRITS to the project, including: (i) the creation of novel datasets for testing technologies (ii) development of an automatic model for categorizing Pagella Politica verdicts to facilitate broader analysis (iii) creation of an automatic model for recognizing textual entailment with exceptional accuracy on the FEVER dataset (iv) assessment using GPT-4 to detecting content treatment style (v) a game to raise awareness about fake news at national events. 
\end{abstract}


\section{Introduction}
The Italian Digital Media Observatory (IDMO) \footnote{\url{https://www.idmo.it/}} project is an  European initiative focused on combating disinformation and fake news. As participants from Rai-CRITS (Centro Ricerche, Innovazione Tecnologica e Sperimentazione, the R\&D\&I of Rai), we have actively contributed to this project. This report provides an overview of the activities we undertook as part of the IDMO project. In particular, our contribution is articulated as follows (contributions of greater significance are marked with one or two asterisks):
\begin{itemize}
    \item Novel datasets on which to test the developed technologies (see section \ref{sec:datasetscreation}).
    \item Two analyses to measure the media's contribution to disinformation:
    \begin{itemize}
        \item An analysis of the impact and influence of mainstream media on Twitter regarding disinformation statements. This analysis did not yield statistically significant findings (see section \ref{subsec:twitterimpat}).
        \item A source-based analysis of Pagella Politica debunking on politicians' statemens, to identify the influence of mainstream media, even indirectly through guest appearances, in the realm of political disinformation discourse. The number of verdicts varied significantly across different media sources, making it challenging to determine which specific media outlets contributed more significantly to disinformation. Among mainstream media with a similar number of verdicts (such as La7 and Rai) and the YouTube platform, Rai appears to the least susceptible to political disinformation. It's important to emphasize that this is only a preliminary result obtained with an average sample size of 20 per source (see section \ref{subsec:politicianstatementmedia}).
    \end{itemize}    
    \item[*] An automatic model designed to categorize Pagella Politica verdicts, for example assigning categories such as ``Exaggeration'' for verdicts like ``Politician X is exaggerating a bit'', or ``True'' for the verdict ``The president is correct". The aim is to facilitate a more comprehensive analysis beyond individual verdicts, enabling answers to broader questions. For instance, it can help identify which parties have exaggerated the most in their statements or which ones have made more omissions. The model has shown that as the training dataset size increases, so does the accuracy on the test set. With only 150 training verdicts, it achieves an accuracy of nearly 85\% (see section \ref{subsubsec:DistilBERTonFever}).  
    \item An analysis was conducted to identify the optimal criterion for inferring similarity between a document and a disinformation statement in retrieval (see section \ref{subsec:documentstatementsimilarity}).
    \item[**] An automatic model aimed at recognizing textual entailment in the context of disinformation statements, notably achieving remarkable results on the challenging FEVER dataset \cite{FEVERPaper} . In particular, the creation of a multilingual \textit{DistilBERT} model has showcased accuracy rates exceeding 90\%, surpassing the performance of algorithms examined in the FEVER paper (see section \ref{subsubsec:DistilBERTonFever}). 
    \item An assessment using GPT-4 to automatically identify textual entailment (see section \ref{subsec:contenttreatementdetection}).
    \item A proposal to harness the capabilities of large language models in order to evaluate text clarity, especially for those composing debunking verdicts (see section \ref{sec:textclaritycheck}).
    \item A game aimed at raising awareness about fake news, presented in various occasions at national events (see section \ref{sec:fakeradar}).
\end{itemize}

As it results from the above listing, more than a self-contained deliverable component addressing a particular function, we oriented our approach towards exploring a set of possible technical supports in the area of misinformation detection and combat, with the idea of shedding light on what technical approaches would be worth being deepened in a later phase in a number of relevant scenarios. 

N.B. All the experiments in the remainder of this paper for which AI models were trained or fine-tuned were conducted using an NVIDIA A100-PCIE-40GB GPU.


\section{Dataset Creation and Annotation}
\label{sec:datasetscreation}

\subsection{Manual Annotatated Datasets}

\begin{figure}[!ht]
    \centering
    \includegraphics[width=0.5\linewidth]{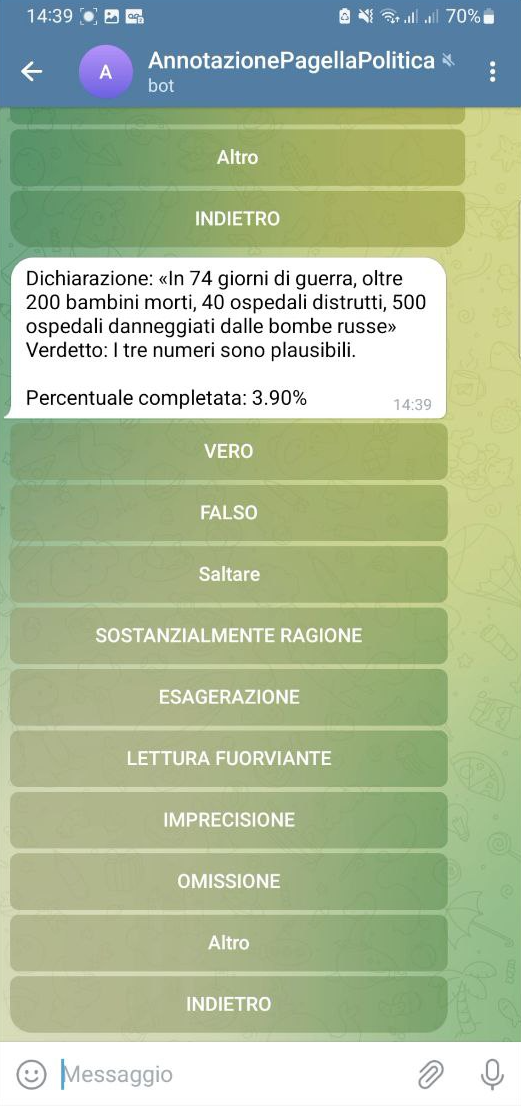}
    \caption{Screenshot depicting the annotation of the PagellaPolitica2 dataset conducted through a Telegram bot for ease of use}
\end{figure}

As part of our involvement in the IDMO project, we were responsible for the creation of several datasets, each tailored to address specific aspects of disinformation detection. These datasets were meticulously annotated to provide a comprehensive understanding of how statements, content treatment, and veracity assessment intertwine in the realm of disinformation.

\subsubsection{Datasets from RAI-CMM and OPEN }
\label{contenttreatmentanalysi}

Our research involved the creation and annotation of two distinct datasets, RAI-CMM-OPEN1 and RAI-CMM-OPEN2. 

\begin{itemize}
\item \textbf{RAI-CMM-OPEN1}: \textit{Text-Statement Similarity}. 
The RAI-CMM-OPEN1 dataset was specifically designed to evaluate the similarity between statements and text, a fundamental aspect of disinformation detection. Our annotation process for this dataset focused on two primary categories:
\begin{itemize}
    \item \textbf{Mention (M):} In this category, we annotated instances where the text included sentences directly related to the statement under consideration.
    \item \textbf{Off-topic (O):} Conversely, we categorized instances where the retrieved resource was unrelated to the statement, i.e., ``Off-topic.''
\end{itemize}

\item \textbf{RAI-CMM-OPEN2}: \textit{Content Treatment Analysis}.
This dataset was specifically designed to analyze how content related to statements is treated by the media.
The annotation process for this dataset encompassed several key strata of classification:
\begin{itemize}
    \item \textbf{Orientation:} Within this stratum, we categorized content into ``Reinforcing'' (indicating support for the statement), ``Confuting'' (signifying opposition to the statement), and ``Neutral'' (indicating a neutral stance toward the statement).
    \item \textbf{Subject:} This stratum involved the classification of content into ``Direct'' (directly affirmed by the publisher reporting the content), ``Indirect'' (reported by third parties, e.g., ``Person X said that...''), and ``Fiction'' (found in fictional sources, such as books).
    \item \textbf{Argumentation:} Within this particular stratum, we gauged the depth of argumentation by categorizing content as either ``Explained'' (explicitly providing reasons for support or opposition) or ``Shallow'' (lacking explicit reasons).
\end{itemize}
\end{itemize}

\subsubsection{Datasets from Pagella Politica }

Our contributions also extended to the incorporation of datasets from Pagella Politica\footnote{\url{https://pagellapolitica.it/fact-checking}}, a renowned source for fact-checking political statements, who was our project partner. It's important to note that annotating these datasets, despite their derivation from website data, required a significant manual effort.

\begin{itemize}
    \item \textbf{PagellaPolitica1:} This dataset focused on the relationship between statements and explanations. It included annotations such as ``Supported'' (supporting the statement), ``Refuted'' (contradicting the statement), and ``Not Enough Info'' (insufficient information to determine support or contradiction).

    \item \textbf{PagellaPolitica2:} In this dataset, we categorized statements' veracity into labels ranging from ``VERO'' (true) to ``FALSO'' (false), with intermediate labels such as ``RAGIONE A METÀ'' (partially true), ``IMPRECISIONE'' (inaccuracy), ``ESAGERAZIONE'' (exaggeration), as well as other labels like ``CONFUSIONE'' (confusion), ``PLAUSIBILE'' (plausible), ``SOSTANZIALMENTE RAGIONE'' (substantially true), ``INFONDATEZZA'' (unfounded), ``MESSAGGIO FUORVIANTE'' (misleading message), ``OMISSIONE'' (omission), ``APPROSSIMAZIONE'' (approximation), ``LETTURA FUORVIANTE'' (misleading reading), ``PRESUPPOSTO SBAGLIATO'' (wrong assumption), and ``TRAVISAMENTO'' (distortion). This comprehensive categorization allows for a nuanced and detailed assessment of fact-checking outcomes.
\end{itemize}

\subsection{Additional Datasets}

In addition to the collection and manual annotation of new datasets, we leveraged existing datasets to further enrich and benchmark the results of our research. These additional datasets include:

\begin{itemize}
    \item \textbf{FEVER \footnote{\url{https://fever.ai/dataset/fever.html}}
:} Originally in English, the FEVER dataset \cite{FEVERPaper} consists of 185,445 claims generated by altering sentences from Wikipedia. These claims are classified as ``Supported'', ``Refuted'', or ``Not Enough Info''. To make this dataset accessible for our research in Italian, we used Google Translator to create an Italian version, called \textit{FEVER-it}. Finally we formed \textit{FEVER-ml}, by the union of the English samples with their Italian translations; it includes all possible claim-text combinations, such as English claim - English text, English claim - Italian text, and so on;
    
    \item \textbf{Ministry of Health:} This dataset, provided by the Ministry of Health  \footnote{\url{https://www.salute.gov.it/portale/nuovocoronavirus/archivioFakeNewsNuovoCoronavirus.jsp}}, addresses COVID-19 misinformation by presenting pairs of false statements and their corresponding true counterparts.
\end{itemize}

\section{Impact of Disinformation Spread through Media}

In our efforts to understand the impact of disinformation spread through various media channels, we conducted an analysis divided into two parts.

\subsection{Twitter Impact Analysis}
\label{subsec:twitterimpat}

The initial part of this analysis, conducted in collaboration with LUISS University, aimed to assess the influence of media sources on Twitter when discussing disinformation statements from the RAI-CMM-OPEN1 and RAI-CMM-OPEN2 datasets. The objective was to determine whether these discussions had a significant impact on Twitter engagement.
LUISS researchers examined the volume of Twitter mentions and interactions related to these sources. The final findings did not indicate a significant deviation from typical engagement patterns. 



\subsection{Politician Statements and Media Influence}
\label{subsec:politicianstatementmedia}
In the second phase of our analysis, we shifted our focus to the PagellaPolitica2 dataset, which contains information about the sources used by politicians when making statements. We examined statements made by politicians, considering only the categories ``VERO'' (true), ``FALSO'' (false), and ``ESAGERAZIONE'' (exaggeration), over a period spanning from February 2022 to July 2023. 

\begin{figure}[H]
    \centering
    \includegraphics[width=0.8\linewidth]{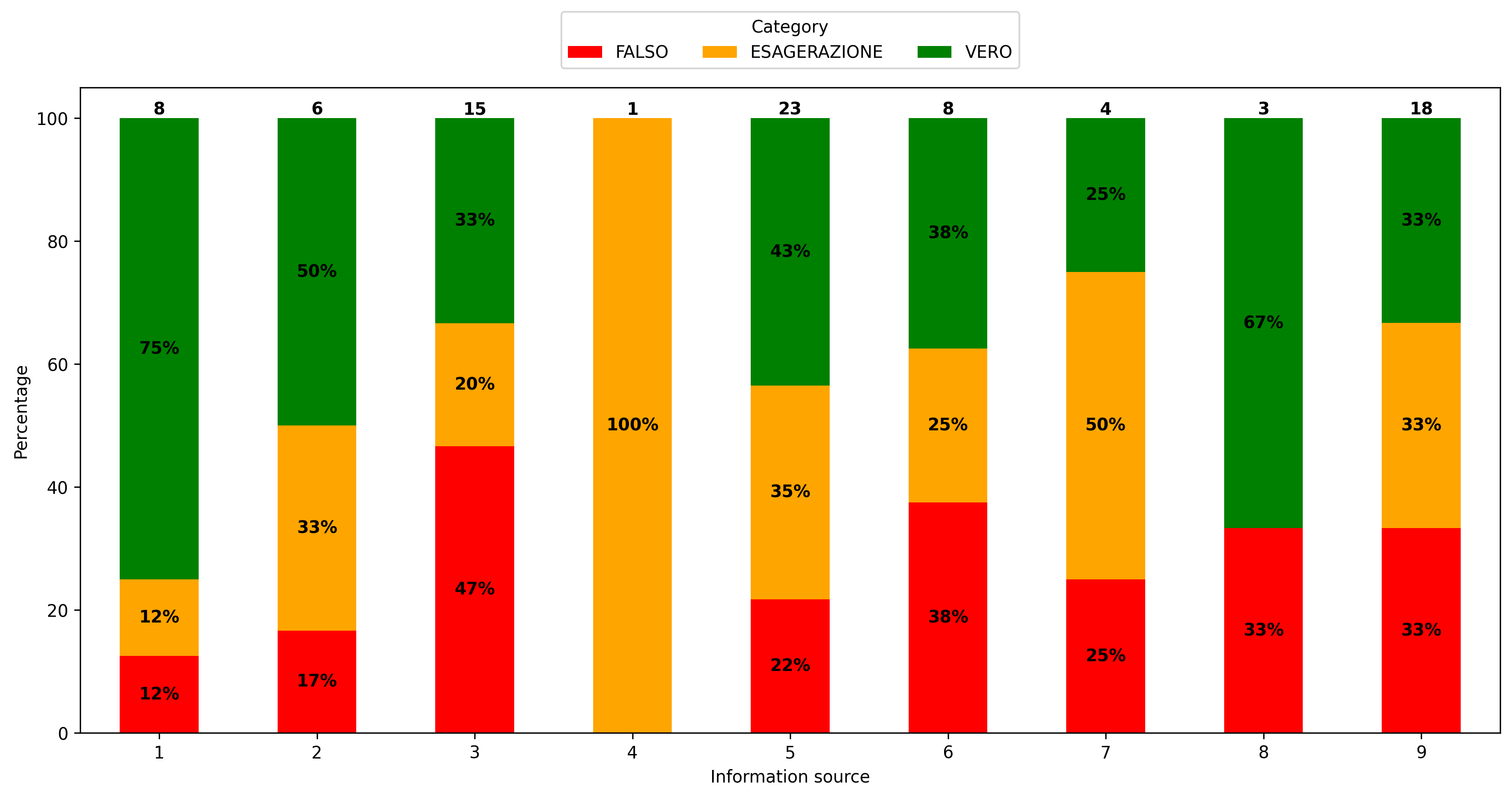}
    \caption{ Distribution of categories by information source}
    \label{fig:evolutionOfCategoriesOverMedia}
\end{figure}

 We analyzed the intertwining between these categories and media sources.
The stacked bar chart in Figure \ref{fig:evolutionOfCategoriesOverMedia} represents how these categories are distributed among different sources of information from January 2022 to the present and provides an overview of the percentages of each evaluation compared to the total evaluations for each source.  Hence this type of chart is useful for examining how different sources report politicians' statements and how such statements are assessed in terms of truthfulness. The numerical values above each bar represent the total count of verdicts for that respective bar. The names of these sources have been anonymized because we lack certainty that the denominator for analysis is common across them and only some political verdicts have been analyzed, not all. This anonymization is crucial to ensure that the assessment remains unbiased and does not misrepresent the reliability of the sources.

Comparing the sources can be challenging, as the verdict counts vary significantly from one source to another, ranging from a minimum of 1 to a maximum of 23.
When considering the information sources with a more substantial number of verdicts, specifically sources 3, 5, and 9, it is evident that source 5 has a marginally lower percentage of false statements compared to the others. Similarly, this source has a slightly higher percentage of true statements and exaggerations. However, given the limited discrepancy and the relatively low number of verdicts considered, we cannot consider these data as representative of a comprehensive analysis.





\section{Usage of Pagella Politica verdicts for further analysis}

\begin{figure}[H]
    \centering
    \includegraphics[width=0.8\linewidth]{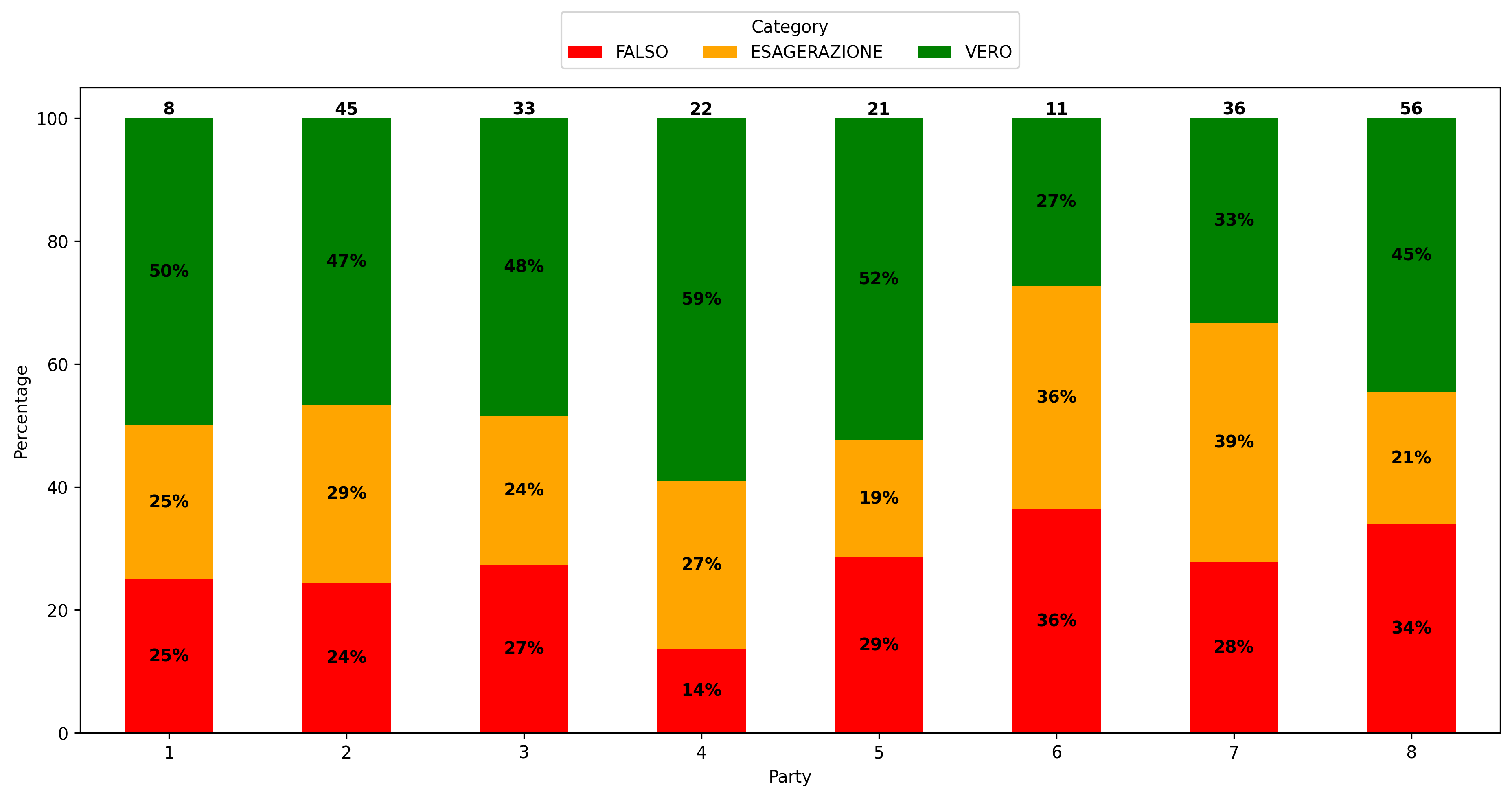}
    \caption{ Distribution of categories by party}
    \label{fig:evolutionOfCategoriesParty}
\end{figure}

In the previous section, we provided a brief analysis of the propagation of fake news and true information through the media, without finding significant evidence. 
However the annotations in the PagellaPolitica2 dataset may have additional validity in conducting a more comprehensive analysis, rather than focusing solely on individual statements,  addressing questions such as: Which political party is most prominent in this category? How do its statements regarding truthfulness change over time? Is there an underlying narrative style for that party? Figure \ref{fig:evolutionOfCategoriesParty} reports some preliminary insights; the parties names have been anonymized because we lack certainty that the denominator for analysis is common across them and only some political verdicts have been analyzed, not all.

It is indeed well-known that the greater the attention to political fake news, the more attention politicians themselves pay to avoiding them.
Future work could explore areas such as tracking the performance of individual politicians over time or developing a tool to help citizens critically evaluate political information through Pagella Politica, addressing the confirmation bias that arises when individuals focus on isolated statements without a global perspective. Wood and Porter, in their study \cite{WoodThomas}, involved over ten thousand individuals by presenting corrections to statements made by political figures. They concluded that ``the evidence for a boomerang effect is much weaker than suggested by previous research. Significantly, citizens seek fact-based information, even when such information contradicts their ideological positions.''

Given the public interest in these analyses, since the previous graphs were only attainable after the extensive effort of manual annotation for verdict categorization, it is intriguing to explore the potential of utilizing annotated data to create an artificial intelligence model capable of inferring categories automatically.

\subsection{Automatically categorize Pagella Politica fact-checking verdicts}

In this section, we describe the experiments conducted to automatically categorize fact-checking verdicts provided by journalists regarding statements made by politicians. 

In particular, we adopted \textit{Camoscio} model \cite{santilli2023camoscio} \footnote{Link to \textit{Camoscio} repository: \url{https://github.com/teelinsan/camoscio}}, an Italian instruction-tuned LLaMA\footnote{\url{https://ai.meta.com/llama/}} based on Stanford Alpaca\footnote{\url{https://github.com/tatsu-lab/stanford_alpaca}} and trained with low-rank adaptation (LoRA, \cite{hu2022lora}). This model is designed to handle natural language processing tasks, including text classification and understanding Italian language nuances.

\begin{figure}[ht]
    \centering
    \includegraphics[width=0.8\textwidth]{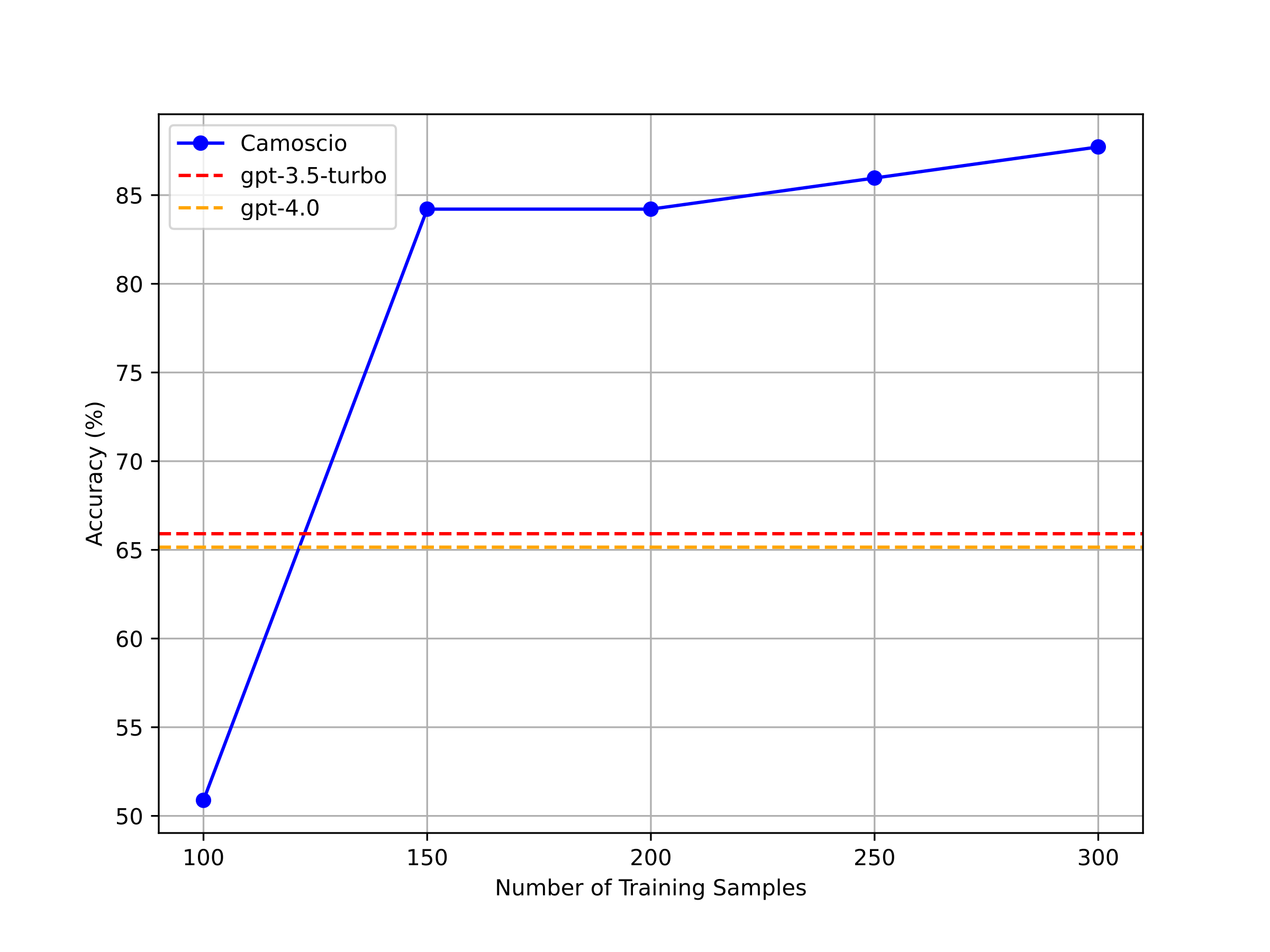}
    \caption{Accuracy vs. Number of Training Samples}
    \label{fig:AccuracySamples}
\end{figure}

We fine-tuned the base \textit{Camoscio} version, incrementally increasing the number of training samples provided to the model. The training samples were ordered chronologically, starting from the earliest available data and progressing towards the most recent; our primary objective was to evaluate how the accuracy of the \textit{Camoscio} model evolved as we increased the number of training samples (see Figure \ref{fig:AccuracySamples}). 

We fine-tuned the base \textit{Camoscio} version, incrementally increasing the number of training samples provided to the model. The training samples were ordered chronologically, starting from the earliest available data and progressing towards the most recent; our primary objective was to evaluate how the accuracy of the \textit{Camoscio} model evolved as we increased the number of training samples (see Figure \ref{fig:AccuracySamples}). The test sample consisted of 57 samples, corresponding to the most recently inputted data by Pagella Politica journalists.

 Remarkably, even with a relatively small dataset of 150 samples, it already achieved a substantial level of accuracy (84\%) in categorizing journalist verdicts regarding political statements.
 As seen in the the figure, we also experimented with GPT-3.5-turbo and GPT-4 without fine-tuning.  The results from these models were noticeably lower, achieving approximately 65\% accuracy.
This is the English translation of the prompt provided to both gpt-3.5-turbo and gpt-4: \textit{This is a verdict regarding the debunking of a politician's claim: \textless VERDICT \textgreater. Based on the text of the verdict, categorize it by returning one of the following labels: true, false, partially true, inaccurate, exaggeration}.

In summary, our experiments indicate that \textit{Camoscio} can effectively categorize journalist verdicts with a high degree of accuracy, even with a limited amount of fine-tuning data, avoiding or reducing the effort to manually categorize them. 

The fine-tuning process took approximately ten minutes due to the limited amount of data.


\section{Document retrieval for debunking}

Document retrieval for debunking involves, at first, retrieving all relevant resources pertaining to a given false claim or news item from a document set (\textbf{Statement-Document Similarity Analysis}, see subsection \ref{subsec:documentstatementsimilarity}).

Secondly it's useful determining how these retrieved documents address the claim or news item. Specifically, they may either support, refute, or maintain a neutral stance towards it (\textbf{Recognizing Textual Entailment}, see subsection \ref{subsec:recognizetextualentailment}). 

In addition, within the realm of editorial journalism, it is valuable to identify the entities within the retrieved documents expressing opinions regarding the claim. These entities could be the journalistic sources themselves, third-party sources (e.g., interviewees) or artworks as books. Furthermore, the initial claim can be supported either superficially by inferring its veracity or falseness or comprehensively by presenting substantial evidence and data to bolster the argument; we call this phase \textbf{Content Treatment Detection} (see subsection \ref{subsec:contenttreatementdetection}).

The subsequent subsections delve into the analysis of natural language processing algorithms to automatically support the previously discussed tasks.

\subsection{Statement-Document Similarity Analysis}
\label{subsec:documentstatementsimilarity}
Our contributions encompassed the evaluation of various natural language processing algorithms' performance in recognizing text-statement similarity, starting from the assumption that a higher similarity means higher relevance of the statement in the text, implying that the statement is somehow treated within the document. This evaluation primarily utilized the RAI-CMM-OPEN1 dataset. We aimed to recognize similarity between statement-document pairs, distinguishing between similar pairs (labeled as $1$) and dissimilar pairs (labeled as $0$). To achieve this, we first divided the dataset into training and testing samples. Key metrics, including Precision, Recall, and F1-Score, were  recorded, providing valuable insights into algorithm effectiveness.
All tested techniques are listed in the following of this section. For each of them indicate document representation, similarity and details.

The \textbf{document representation} indicates how the two texts to be compared (statement and document) were represented, i.e. either numerically through embeddings or using sets of named entities. \textbf{Similarity} denotes the specific similarity measures adopted between the two representations, such as cosine distance or the negative of Euclidean distance for numerical vectors and set-based similarities for entities. The \textbf{details} provide additional information on how the representations were obtained, including variations in TFIDF parameters through grid search or the choice of models for BERT.

\begin{enumerate}
    \item 
    \begin{itemize}
        \item \textbf{Documents Representation}: TFIDF embeddings of the statement and the retrieved resource
        \item \textbf{Similarity}: Cosine similarity, - Euclidean distance
        \item \textbf{Details}: Grid search parameters:
        \begin{itemize}
            \item ``n\_features'':$[20,50,100,200,500,1000]$
            \item ``ngram\_range'':$[(1,1),(2,2),(3,3),(1,3),(1,2),(2,3)]$
            \item ``max\_df'':$[0.7,0.8,0.9,1.0]$
            \item ``min\_df'':$[1,3,5,7,10]$
        \end{itemize}
        These parameters are derived from scikit-learn (sklearn) \footnote{\url{https://scikit-learn.org/stable/modules/generated/sklearn.feature_extraction.text.TfidfVectorizer.html}}.
    \end{itemize}
    
    \item 
    \begin{itemize}
        \item \textbf{Document Representations}: BERT embeddings of the statement and the retrieved resource
        \item \textbf{Similarity}: Cosine similarity, - Euclidean distance
        \item \textbf{Details}: Bert models (from HuggingFace \footnote{\url{https://huggingface.co}}) : \textit{bert-base-italian-xxl-cased, bert-base-italian-cased, bert-base-italian-xxl-uncased, bert-base-italian-xxl-cased, distiluse-base-multilingual-cased}
    \end{itemize}
    
    \item 
    \begin{itemize}
        \item \textbf{Document Representation}: BERT embeddings of the statement and the retrieved resource translating both texts in English
        \item \textbf{Similarity}: Cosine similarity, - Euclidean distance
        \item \textbf{Details}: Bert models (from HuggingFace) : \textit{paraphrase-multilingual-mpnet-base-v2, distiluse-base-multilingual-cased-v1, distiluse-base-multilingual-cased-v2, paraphrase-multilingual-MiniLM-L12-v2}
    \end{itemize}
    
    \item 
    \begin{itemize}
        \item \textbf{Document Representations}: BERT embeddings of the statement and each of the sentences in the retrieved resource
        \item \textbf{Similarity}: Cosine similarity (the highest value between each statement-sentence pair), - Euclidean distance (the highest value between each statement-sentence pair)
        \item \textbf{Details}: Bert models (from HuggingFace) : \textit{bert-base-italian-xxl-cased, bert-base-italian-cased, bert-base-italian-xxl-uncased, bert-base-italian-xxl-cased, distiluse-base-multilingual-cased}
    \end{itemize}
    
    \item
    \begin{itemize}
        \item \textbf{Document Representations}: BERT embeddings of the statement and each of the sentences in the retrieved resource translating each text in English with Google Translate
        \item \textbf{Similarity}: Cosine similarity (the highest value between each statement-sentence pair), - Euclidean distance (the highest value between each statement-sentence pair)
        \item \textbf{Details}: Bert models (from HuggingFace) : \textit{paraphrase-multilingual-mpnet-base-v2, distiluse-base-multilingual-cased-v1, distiluse-base-multilingual-cased-v2, paraphrase-multilingual-MiniLM-L12-v2}
    \end{itemize}
    
    \item
    \begin{itemize}
        \item \textbf{Document Representations}: Named entities sets extracted with Spacy
        \item \textbf{Similarity}: Jaccard similarity, Dice similarity, Overlap similarity, Normalized weighted intersection, Cosine sets similarity, Union
    \end{itemize}
    
    \item
    \begin{itemize}
        \item \textbf{Document Representations}: Named entities sets extracted with Spacy from the texts translated in English with Google Translate
        \item \textbf{Similarity}: Jaccard similarity, Dice similarity, Overlap similarity, Normalized weighted intersection, Cosine sets similarity, Union
    \end{itemize}
\end{enumerate}

The  technique that reaches the highest F1-score (0.43)  was TFIDF (Term Frequency-Inverse Document Frequency) using euclidean distance and the following set of parameters: \{``n\_features'': 50, ``ngram\_range'': (2, 3), ``max\_df'': 0.7, ``min\_df'': 1\}.

However, these results were not satisfactory, prompting us to explore alternative approaches. We utilized the similarity scores generated by the previously described techniques (TFIDF, BERT, named entitites) as input vectors for machine learning algorithms, as shown in the table below. These algorithms served as an ensemble and were trained on the training set. The ensemble approach significantly improved performance, achieving an F1-Score of 0.97.
It should be noted that within the ensemble, we evaluated the features that had the greatest impact, and it appears that the combination of TF-IDF scores obtained with different parameters had the most significant influence on the output. The contribution of BERT was indeed marginal, and that of named entities was negligible.


\begin{table}[h]
\centering
\small
\caption{Performance Metrics for Ensemble Algorithms in Statement-Document Similarity task}
\vspace{0.33cm}
\label{tab:performance-metrics}
\begin{tabular}{lccc}
\toprule
\textbf{Algorithm} & \textbf{Precision} & \textbf{Recall} & \textbf{F1-Score} \\
\midrule
\textit{SVC} & 1 & 0.95 & 0.97 \\ 
\textit{KNeighborsClassifier} & 0.95 & 0.98 & 0.96 \\ 
\textit{GaussianProcessClassifier} & 0.84 & 0.86 & 0.85 \\ 
\textit{RandomForestClassifier} & 0.83 & 0.79 & 0.80 \\ 
\textit{DecisionTreeClassifier} & 0.84 & 0.76 & 0.80 \\ 
\textit{KNeighborsClassifier} & 0.70 & 0.74 & 0.72 \\ 
\textit{DecisionTreeClassifier} & 0.76 & 0.62 & 0.68 \\ 
\textit{GaussianNB} & 0.28 & 0.98 & 0.44 \\ 
\textit{best TFIDF (not ensemble)} & 0.27 & 1 & 0.43 \\
\bottomrule
\end{tabular}
\end{table}

While the performance is very high, the downside of the ensemble is that it requires a large number of features as input, which may result in slower response times  for a real-time retrieval system. retrieval system. For this reason, currently, we are actively exploring LangChain\footnote{\url{https://python.langchain.com/}} as a potential solution to enhance our retrieval system. LangChain is a framework designed for developing applications powered by language models, and it seamlessly integrates with FAISS\footnote{\url{https://ai.meta.com/tools/faiss/}}, a powerful similarity search library. Our focus within LangChain lies in the realm of document chunking and employing various embeddings to achieve optimal results for a faster retrieval system.




\subsection{Recognizing Textual Entailment}
\label{subsec:recognizetextualentailment}

Recognizing Textual Entailment (RTE), also known as Natural Language Inference (NLI), plays a pivotal role in our research. In the realm of Natural Language Processing (NLP), RTE involves determining if the content of one text fragment logically implies the content of another. Specifically, we say ``text entails hypothesis'' ($t \rightarrow h$) if, under typical circumstances, a human reader would reasonably conclude that statement $h$ is highly likely to be true based on the content of text $t$. 
More simply RTE involves determining one of the following classes for each pair of a textual claim and hypothesis: ``Supported'', ``Refuted'', or ``Not Enough Info''.

\subsubsection{\textit{DistilBERT} fine-tuning on FEVER dataset}
\label{subsubsec:DistilBERTonFever}
In our first experiment, we utilized the FEVER (Fact Extraction and Verification\footnote{\url{https://fever.ai/dataset/fever.html}}) dataset, a widely recognized benchmark for RTE tasks. The dataset contains textual claims paired with corresponding hypotheses, and our goal was to determine whether the hypothesis could be logically inferred from the text.

For this experiment, we fine-tuned \textit{DistilBERT} \cite{DistilBERTPaper}, a compact version of the BERT \cite{bert} model; in particular  we opted for the ``distilbert-base-multilingual-cased'' \footnote{\url{https://huggingface.co/distilbert-base-multilingual-cased}} variant, which is pretrained on a diverse set of languages, making it suitable for our multilingual evaluation.
The \textit{DistilBERT} models took approximately 1 hour for the training.

In addition we also trained \textit{Camoscio} on the same dataset. However, the training for \textit{Camoscio} was conducted with a smaller subset of the data, specifically 18,000 samples out of the available 131,583 samples, and for only one training epoch. This approach was chosen to accommodate the longer training times required for \textit{Camoscio}, which exceeded 24 hours per epoch.

To provide a comprehensive benchmark, we compared the fine-tuned \textit{DistilBERT}'s performance to  algorithms adopted in \cite{FEVERPaper} for the same dataset using the random sampling (RANDOMS) approach (see Table \ref{tab:model-accuracy-FEVER-benchmark}).

\begin{table}[H]
    \centering
    \caption{\textit{DistilBERT} Accuracy (\%) againist competitors in Recognizing Textual Entailment on FEVER dataset}
    \vspace{0.33cm}
    \label{tab:model-accuracy-FEVER-benchmark}
    \begin{tabular}{lccc}
        \toprule
        \textbf{Model} & \textbf{Accuracy (\%)}  \\
        \midrule
        \textit{MLP} & 65.13 \\
        \textit{DA} & 80.82 \\
        \textit{DistilBERT} & 89.51 & \\
        \textit{Camoscio} & 87.21  \\
        \bottomrule
    \end{tabular}
\end{table}

Remarkably, our \textit{DistilBERT} model achieved an accuracy of 89.51\%,  outperforming the competing models. Notably, \textit{Camoscio} also attained an accuracy very close to that of \textit{DistilBERT}, and there is a strong possibility that through extended fine-tuning on the complete dataset, rather than just the initial 18,000 samples, it may outperform \textit{DistilBERT}.

To assess the RTE task in Italian and multilingual settings, we fine-tuned other models on FEVER-it and FEVER-ml. For all three dataset versions, we followed the split strategy between training and test data as presented in the FEVER paper. Additionally, we created four more splits, resulting in five datasets in total. Each of the FEVER and FEVER-it datasets contains 9,893 test samples and 131,583 training samples. In contrast, the FEVER-ml dataset, which combines English claims with texts in both English and Italian, contains four times the number of samples, reflecting all possible claim-text combinations, such as English claim - English text, English claim - Italian text, etc.

Table \ref{tab:distilbert-folds} presents the results for all five folds for each dataset, calculated with all models; it also includes the results of models trained on one dataset and evaluated on another.
From the results, the following findings emerge:

\begin{itemize}

    \item Models trained on the FEVER dataset (English language) show high accuracy when evaluated on the same dataset. Specifically, they achieve accuracies ranging from 89.51\% to 99.70\% across different folds, with an average accuracy of 97.58\% and a standard deviation of 4.04\%. The performance decreases for the other datasets since the models reach  average accuracies of approximately 81.65\% (s.t.d 2.92\%) on FEVER-it and 83.71\% (s.t.d. 3.38\%) on FEVER-ml. This is an interesting finding as it suggests that \textit{the FEVER models are impacted by the language change directly by approximately 10 percentage points due to the use of a multilingual version of \textit{DistilBERT}.}

    \item Models trained on FEVER-it show high accuracy when evaluated on the same dataset, achieving accuracies ranging from 87.94\% to 99.52\% across different folds, with an average accuracy of 97.10\% (s.t.d 4.58\%). These performances are comparable to those achieved by the FEVER models on the FEVER dataset, indicating that there is no noticeable difference in performance between the two datasets when tested on the training dataset. Furthermore, the models trained on FEVER-it experience a smaller decrease in accuracy when evaluated on other datasets compared to the decrease observed in the FEVER models, achieving an average accuracy of 87.7\% (s.t.d 1.51\%) on FEVER and of 89.44\% (s.t.d. 2.8\%) on FEVER-ml. For the latter, the performances are higher than those achieved with the FEVER models.

    \item Models trained on FEVER-ml show high accuracy when evaluated on the same dataset, achieving an average accuracy of 87.9\% (s.t.d. 4.77\%), only slightly higher compared to that obtained with FEVER-it models on the same dataset (87.7\%, s.t.d 1.51\%). It indicates the FEVER-it models' ability to generalize across both languages. This is likely because the original \textit{DistilBERT} model is multilingual, and the FEVER-it dataset is an automatic translation of an English dataset using Google Translate, preserving an English language structure.
    Models trained on FEVER-ml exhibit slightly higher accuracy on the other datasets, 90.89\% (s.t.d. 4.45\%) on FEVER and 89.82\% (s.t.d. 4.85\%) on FEVER-it. It's worth noting that the FEVER-ml test set also includes statement-text pairs in both Italian and English, and vice versa. Therefore, it's not a simple combination of the other two datasets, and these additional pairs may be slightly more challenging to predict.
\end{itemize}

We can conclude that the best performances for a dataset are consistently achieved by models trained on the same dataset, with the exception of FEVER-it when evaluated on the FEVER-ml dataset, where it reaches comparable accuracies to those obtained by models trained with FEVER-ml. The FEVER-it models  achieve the highest average results across all three datasets. 


\begin{table}
    \centering
    \caption{\textit{DistilBERT} models accuracies (\%) on each fold in recognizing textual entailment on FEVER dataset}
    \vspace{0.33cm}
    \label{tab:distilbert-folds}
    \begin{tabular}{cccc}
    \toprule
    \textbf{Test Dataset} & \textbf{Training Dataset} & \textbf{Fold} & \textbf{Accuracy (\%)} \\
    \midrule
    \multirow{15}{*}{\textbf{FEVER}} & \multirow{5}{*}{\textbf{FEVER}} & 1 & 89.51 \\
    & & 2 & 99.70 \\
    & & 3 & 99.66 \\
    & & 4 & 99.37 \\
    & & 5 & 99.64 \\
    \cmidrule{2-4}
    & \multirow{5}{*}{\textbf{FEVER-it}} & 1 & 81.84 \\
    & & 2 & 75.96 \\
    & & 3 & 83.85 \\
    & & 4 & 83.25 \\
    & & 5 & 83.36 \\
    \cmidrule{2-4}
    & \multirow{5}{*}{\textbf{FEVER-ml}} & 1 & 78.49 \\
    & & 2 & 80.92 \\
    & & 3 & 86.87 \\
    & & 4 & 86.45 \\
    & & 5 & 85.83 \\
    \midrule
    \multirow{15}{*}{\textbf{FEVER-it}} & \multirow{5}{*}{\textbf{FEVER}} & 1 & 86.86 \\
    & & 2 & 90.43 \\
    & & 3 & 85.95 \\
    & & 4 & 87.96 \\
    & & 5 & 87.30 \\
    \cmidrule{2-4}
    & \multirow{5}{*}{\textbf{FEVER-it}} & 1 & 87.94 \\
    & & 2 & 99.42 \\
    & & 3 & 99.38 \\
    & & 4 & 99.52 \\
    & & 5 & 99.25 \\
    \cmidrule{2-4}
    & \multirow{5}{*}{\textbf{FEVER-ml}} & 1 & 84.04 \\
    & & 2 & 91.71 \\
    & & 3 & 89.50 \\
    & & 4 & 91.24 \\
    & & 5 & 90.70 \\
    \midrule
    \multirow{15}{*}{\textbf{FEVER-ml}} & \multirow{5}{*}{\textbf{FEVER}} & 1 & 95.27 \\
    & & 2 & 94.61 \\
    & & 3 & 92.62 \\
    & & 4 & 88.70 \\
    & & 5 & 83.25 \\
    \cmidrule{2-4}
    & \multirow{5}{*}{\textbf{FEVER}-it} & 1 &  94.54 \\
    & & 2 & 93.57 \\
    & & 3 & 91.79 \\
    & & 4 & 87.94 \\
    & & 5 & 81.24 \\
    \cmidrule{2-4}
    & \multirow{5}{*}{\textbf{FEVER}-ml} & 1 &  94.77 \\
    & & 2 & 90.10 \\
    & & 3 & 88.17 \\
    & & 4 & 86.25 \\
    & & 5 & 80.23 \\
    \bottomrule
    \end{tabular}
\end{table}

\subsubsection{Results on Italian fake-news datasets}

\begin{table}[!ht]
    \centering
    \caption{Dimensionality of Datasets implied in Recongizing Textual Entailment task}
    \vspace{0.33cm}
    \label{stast-datasets}
    \begin{tabular}{lccc}
        \toprule
        \textbf{Dataset Name} & \textbf{Training Samples} & \textbf{Test Samples} \\
        \midrule
        \textbf{FEVER-it} & 131583 & 9893 \\
        \textbf{FEVER-it-small} & 198 & 99 \\
        \midrule
        \textbf{PagellaPolitica1} & 200 & 66 \\
        \midrule
        \textbf{FakeNewsMinistero} & - & 77 \\
        \bottomrule
    \end{tabular}
\end{table}

Furthermore, we conducted an evaluation of \textit{DistilBERT} on the PagellaPolitica1 and Fake News Ministry datasets. To compare the results with FEVER-it and fine-tune models that combined the annotation styles of multiple datasets, we created a smaller version of FEVER-it called FEVER-it-small. This subset of the dataset is also useful for gaining an understanding of the results that GPT-3.5-turbo, GPT-4, and H2O's large language models \footnote{\url{https://gpt.h2o.ai/}} have on the FEVER-it data.Since obtaining responses from these models requires longer processing times, as we rely on third parties, using the entire FEVER-it dataset is not feasible. OpenAI does not release its models, and H2O's models require computational resources, specifically GPUs, which we do not have access to.

\subsubsubsection{Datasets analysis and hetereogeneity}

The FEVER dataset primarily revolves around general world knowledge, covering a wide array of topics. It  can include a broad spectrum of statements. In contrast, the PagellaPolitica1 dataset specializes in fact-checking claims related to politics and political figures in Italy. It is focused on statements and assertions made within the context of Italian politics, making it domain-specific. The Fake News Ministry dataset, similarly, is dedicated to detecting and debunking fake news, misinformation, and disinformation related to the COVID-19 pandemic.

From a quantitative and annotation effort perspective, the FEVER dataset stands out with a significantly larger number of data points, as evident in Table \ref{stast-datasets}. These quantities are incomparable to those of the other datasets, and FEVER also benefits from a higher number of annotators. 

On the other hand, when we delve into a more qualitative analysis of the different annotation meanings across datasets, we extracted five samples from each dataset, as outlined in Tables \ref{samplesFNM}, \ref{samplesPagellaPolitica1}, and \ref{samplesFEVER}.

In FEVER dataset, the majority of supporting texts are generally clear and comprehensive. For example, in the statement ``Stephen Colbert is the host of The Late Show on CBS'' the supporting text provides specific details, indicating that Colbert has been the host of ``The Late Show with Stephen Colbert'', a late-night television talk show on CBS, since September 8, 2015. However, there are two supporting texts highlighted in yellow and light yellow where  the subjects involved are not explicitly mentioned (higher level of yellow indicates a lower level of subject explicitness). For instance, one of the yellow-highlighted annotations claims that ``Sebastian Stan acted in Political Animals'', but the supporting text uses pronouns (``His'') without mentioning  Sebastian Stan. 
Similarly, in the text, ``Born in New Orleans, Rice spent most of her early years there before moving to Texas and later to San Francisco'', the subject's full name, "Anne Rice", is not explicitly mentioned. Instead, it relies on implicit references to ``Rice'' and ``her''.
These ambiguities could pose a challenge in fully comprehending the assertion and evaluating its accuracy without prior knowledge. However it's important to note that this dataset was originally created within a context where only Wikipedia documents deemed relevant based on specific queries, i.e., the claims, were initially filtered. The determination of textual entailment was subsequently made. Consequently, the provided claim-text pair may not always suffice to infer an accurate verdict particularly in cases involving ambiguous pronouns, as it assumes familiarity with the referenced document. 

Unlike the FEVER dataset, in PagellaPolitica1, the primary point of concern lies not so much in the supporting texts but in the claims themselves. While the supporting texts are generally helpful in explaining the verdict and are in line with it, some claims have  limitations.
For example, consider the claim: ``The water network wastes about 40 percent of our water.'' This claim lacks specificity regarding the time it was made. Without a reference date or a specific period, it becomes challenging to establish its accuracy.
Another example pertains to the lack of clarity in the involved subjects. A statement like ``We don't spend money to buy weapons that we send to the Ukrainians'' does not specify who the subject ``we'' refers to; therefore, the issue of ambiguous pronouns primarily manifests within the claims themselves.
Hence, for this dataset, in addition to the issue of implicit texts, there is also a lack of temporal references (orange annotations).
However, it's important to note that these limitations within the claims should also be considered within a specific context, similar to the case of FEVER and the context of Wikipedia: on the Pagella Politica website \footnote{Link to Pagella Politica Fact checking website: \url{https://pagellapolitica.it/fact-checking}}, each claim is accompanied not only by the claim itself and the supporting text but also by information about who made the claim and when it was made.

The supporting texts in Ministery of Health dataset are notably more extensive and detailed compared to those in other datasets.  For instance, in the claim regarding shoes as a transmission vector for the virus, the supporting text not only refutes the claim but also offers practical advice to mitigate potential risks, emphasizing its rich contextual content.
These detailed supporting texts are valuable as they present comprehensive explanations, scientific evidence, and practical recommendations. 
Similarly, the claims appear to be relatively clear and straightforward. They do not pose significant challenges in terms of clarity or ambiguity.

\subsubsubsection{Experimental results}


Table \ref{tab:prompts} shows the prompts given to Large Language Models. The prompts are provided in Italian due to the Italian-language datasets, along with their English translations for clarity.
The second prompt is a simplified version of the first one. In the first prompt, the expected response format is provided, and it was used for all non-finetuned models (except for one version of \textit{Camoscio}, which was also tested with the second prompt). \textit{DistilBERT}, being the only non-large language model (LLM) in Table \ref{tab:model-accuracy}, does not require a prompt but only the input pair.

\begin{table}[!ht]
    \centering
    \caption{Prompts given to the large language models for Recognizing Textual Entailment}
    \vspace{0.33cm}
    \label{tab:prompts}
    \renewcommand{\arraystretch}{1.2}
    \begin{tabularx}{\textwidth}{cXX}
        \toprule
        \textbf{Prompt} & \textbf{Italian} & \textbf{English} \\
        \midrule
        \textbf{1} & FRASE: \textless frase\textgreater & PHRASE: \textless phrase\textgreater \\
         & TESTO: \textless testo\textgreater & TEXT: \textless text\textgreater \\
         & Data la seguente coppia FRASE e TESTO, dimmi se il TESTO supporta la FRASE, la confuta o non c'è abbastanza informazione per dirlo rispondendo con 0 (supporta), 1 (confuta), 2 (not enough info). Esempio di output: 0 (supporta) il testo supporta la frase perché... Esempio di output: 1 (confuta) il testo confuta la frase perché... Esempio di output: 2 (not enough info) il testo... & Given the following pair of PHRASE and TEXT, tell me if the TEXT supports the PHRASE, refutes it, or there is not enough information to determine by responding with 0 (supports), 1 (refutes), 2 (not enough info). Example output: 0 (supports) the text supports the phrase because... Example output: 1 (refutes) the text refutes the phrase because... Example output: 2 (not enough info) the text... \\ \midrule
        \textbf{2} & FRASE: \textless frase\textgreater & PHRASE: \textless phrase\textgreater \\
         & TESTO: \textless testo\textgreater & TEXT: \textless text\textgreater \\
         & Data la seguente coppia FRASE e TESTO, dimmi se il TESTO supporta la FRASE, la confuta o non c'è abbastanza informazione per dirlo rispondendo con 0 (supporta), 1 (confuta), 2 (not enough info). & Given the following pair of PHRASE and TEXT, tell me if the TEXT supports the PHRASE, refutes it, or there is not enough information to determine by responding with 0 (supports), 1 (refutes), 2 (not enough info). \\
        \bottomrule
    \end{tabularx}
    \smallskip
\end{table}

Table \ref{tab:model-accuracy} displays the results obtained on three distinct datasets in terms of accuracy. The models are categorized into two groups: non-fine-tuned models and fine-tuned models. For PagellaPolitica1, we also calculated the \textit{Balanced accuracy (Bal Acc)} metric to assess the performance that that takes into account the effect of class imbalances in the data.  The ``Average'' column in the table represents the average of \textit{Balanced accuracy} for PagellaPolitica1 and \textit{Accuracy} for FEVER-it-small and FakeNewsMinistero, providing a summary measure of model performance across the different datasets.

Notably, the fine-tuned models include models like \textit{Camoscio} and Distillbert and in parentheses next to each, you'll find the datasets on which they were fine-tuned. If there is a comma, it signifies that the model was initially trained on one dataset and subsequently on another. If you see union symbol it indicates that the training data was created by combining both datasets and then fine-tuning the model once.
It's worth noting that there are no fine-tuned models with the Ministry of Health, as this dataset contains only the ``Refuted'' label. Consequently, a model would learn to predict the same class consistently. Therefore, this dataset serves as a valuable test dataset but is not useful for fine-tuning.

\begin{table}[!ht]
    \centering
    \caption{Accuracy (\%) and Balanced Accuracy (\%) of many different models in Recognizing Textual Entailment on three different small Italian datasets}
    \vspace{0.33cm}
    \label{tab:model-accuracy}
    \renewcommand{\arraystretch}{1.2} 
    \begin{tabular}{p{4cm}lcccc}
        \toprule
        \textbf{Model} & \multicolumn{4}{c}{\textbf{Dataset}} & \textbf{Average} \\ 
        &  \multicolumn{2}{c}{\textbf{PagellaPolitica1}}
        & \textbf{FEVER-it-small} & \textbf{MinistryHealth} \\
        \midrule
        & \textit{Acc} & \textit{Bal Acc} & \textit{Acc} & \textit{Acc} \\
        \midrule
        \textbf{Non-fine-tuned models} & & & & & \\
        \textit{h2ogpt-4096-llama2-70b} & 42.42 & 40.00 & 67.68 & 74.03 & 60.57 \\
        \textit{h2ogpt-4096-llama2-13b} & 39.39 & 50.48 & 42.42 & 32.47 & 41.79 \\
        \textit{h2ogpt-4096-llama2-7b} & 36.36 & 53.33 & 35.35 & 1.30 & 30.00 \\
        \textit{h2ogpt-gm-oasst1-en-2048-falcon-40b-v2} & 39.39 & 50.47 & 42.42 & 16.88 & 36.59 \\
        \textit{h2ogpt-falcon-180B} & 63.64 & 61.90 & 69.70 & 77.92 & 69.84 \\
        \textit{gpt-3.5-turbo} & 65.15  & 50.37 & 78.79 & 80.52 & 69.89 \\
        \textit{gpt-4} & \textbf{68.18} & \textbf{74.13} & \textbf{88.89} & 71.43 & \textbf{78.15} \\
        \textit{Camoscio (prompt 1)} & 12.12 & 39.10 & 47.47 & 0.00 & 28.86 \\ 
        \textit{Camoscio (prompt 2)} & 1.51 & 1.11 & 34.34 & 2.60 & 12.68 \\ 
        \midrule
        \textbf{Fine-tuned models} & & & & & \\
        \textit{Camoscio (FEVER-it)} & 9.09 & 11.59 & 80.80 & 29.87 & 40.75 \\
        \textit{Camoscio (FEVER-it-small)} & 57.58 & 48.73 & 39.39 & 64.94 & 51.02 \\
        \textit{Camoscio (PagellaPolitica1)} & 66.67 & 51.43 & 37.37 & 90.09 & 59.63 \\
        \textit{Camoscio (FEVER-it, PagellaPolitica1)} & 48.48 & 49.52 & 85.86 & 27.27 & 54.22 \\
        \textit{Camoscio (PagellaPolitica1, FEVER-it)} & 1.51 &  1.11 & 34.34 & 2.30 & 12.58 \\
        \textit{Camoscio (FEVER-it-small, PagellaPolitica1)} & 59.09 & 47.30 & 34.35 & 70.13 & 50.59 \\
        \textit{Camoscio (PagellaPolitica1, FEVER-it-small)} & 22.73 & 28.73 & 46.46 & 22.08 & 32.42 \\
        \textit{Camoscio (PagellaPolitica1 $\cup$ FEVER-it-small)} & 46.97 & 38.57 & 40.40 & 24.68 & 34.55 \\
        \textit{Distillbert (FEVER-it)} & 15.15 & 20.00 & 82.83 & 9.09 & 37.31 \\
        \textit{Distillbert (FEVER-it-small)} & 42.42 & 36.19 & 48.48 & 42.86 & 42.51 \\
        \textit{Distillbert (PagellaPolitica1)} & 63.63 & 51.75 & 33.33 & 93.51 & 59.53 \\
        \textit{Distillbert (FEVER-it, PagellaPolitica1)} & 56.06 & 57.62 & 82.83 & 53.24 & 64.56 \\
        \textit{Distillbert (PagellaPolitica1, FEVER-it)} & 36.36 & 36.37 & 84.85 & 14.29 & 45.17 \\
        \textit{Distillbert (FEVER-it-small, PagellaPolitica1)} & 59.09 & 54.76 & 42.42 & 87.01 & 61.40 \\
        \textit{Distillbert (PagellaPolitica1, FEVER-it-small)} & 50.00 & 51.90 & 53.53 & 45.45 & 50.29 \\
        \textit{Distillbert (PagellaPolitica1 $\cup$ FEVER-it-small)} & 66.67 & 56.50 & 51.52 & \textbf{93.51} & 67.18 \\
        \bottomrule
    \end{tabular}
\end{table}

None of the fine-tuned models managed to surpass gpt-4 in terms of overall performance, likely due to the limited training samples.  However, on individual datasets, \textit{DistilBERT (PagellaPolitica1 $\cup$ FEVER-it-small)} outperformed GPT-4 on the Ministry Health dataset, and this model stands out as the best performer among the fine-tuned models.

The results achieved by GPT-4 on diverse datasets are still promising, and it could be an interesting avenue for future work to explore whether these results can be further improved through fine-tuning gpt-4 model. 
Furthermore, it's worth noting that, on the small subset of the FEVER-it dataset, GPT-4 performs slightly better than the finetuned \textit{DistilBERT} and \textit{Camoscio} models for FEVER-it, even though \textit{DistilBERT} requires significantly less computational power and can even run on a CPU. Additionally, the \textit{DistilBERT} model is readily available.

One aspect that appears less clear and warrants further analysis is why GPT-4  achieves the lowest score on the Ministry dataset, despite the more detailed explanations provided for debunking statements. Further investigation is needed to understand this phenomenon and any potential factors contributing to GPT-4's performance on this particular dataset.

In the next section, given the complexity of the Content Treatment Statement task, we have only presented preliminary results obtained with GPT-4.

\subsection{Content Treatment Detection}
\label{subsec:contenttreatementdetection}

\begin{table}[!ht]
    \centering
    \caption{Accuracy (\%) and Balanced Accuracy (\%) for Content Treatment Detection on RAI-CMM-OPEN2 dataset}
    \vspace{0.33cm}
    \label{tab:model-accuracy-RAI-CMM-OPEN2}
    \begin{tabular}{lccccccc}
        \toprule 
        \textbf{Model} & \multicolumn{2}{c}{\textbf{Orientation}} & \multicolumn{2}{c}{\textbf{Subject}} & \multicolumn{2}{c}{\textbf{Argumentation}} & {\textbf{All}} \\
        \cmidrule(lr){2-3}
        \cmidrule(lr){4-5}
        \cmidrule(lr){6-7}
        \cmidrule(lr){8-8}
        & \textit{Acc} & \textit{Bal Acc} & \textit{Acc} & \textit{Bal Acc} & \textit{Acc} & \textit{Bal Acc} & {\textit{Acc}} \\
        \midrule
        \textit{gpt-4} & 47.25 & 55.77 & 89.01 & 52.74 & 93.55 & 57.76 & 32.26  \\
        \bottomrule
    \end{tabular}
\end{table}

\begin{figure}[ht]
    \centering
    \begin{subfigure}{0.4\textwidth}
        \centering
        \includegraphics[width=\textwidth]{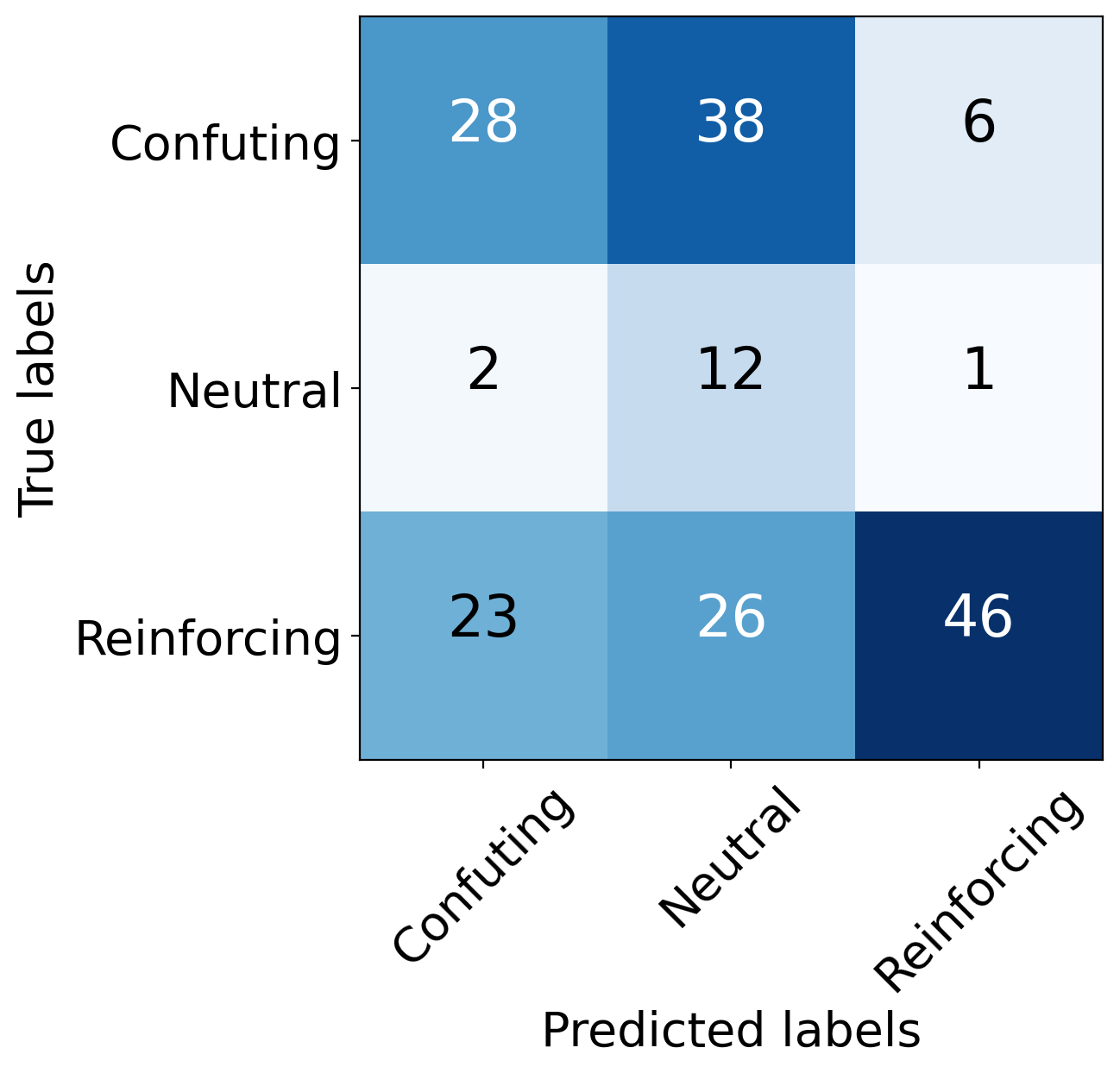}
        \caption{Orientation}
        \label{fig:confusion-matrix-ori}
    \end{subfigure}
        \hspace{1cm} 
    \begin{subfigure}{0.4\textwidth}
        \centering
        \includegraphics[width=\textwidth]{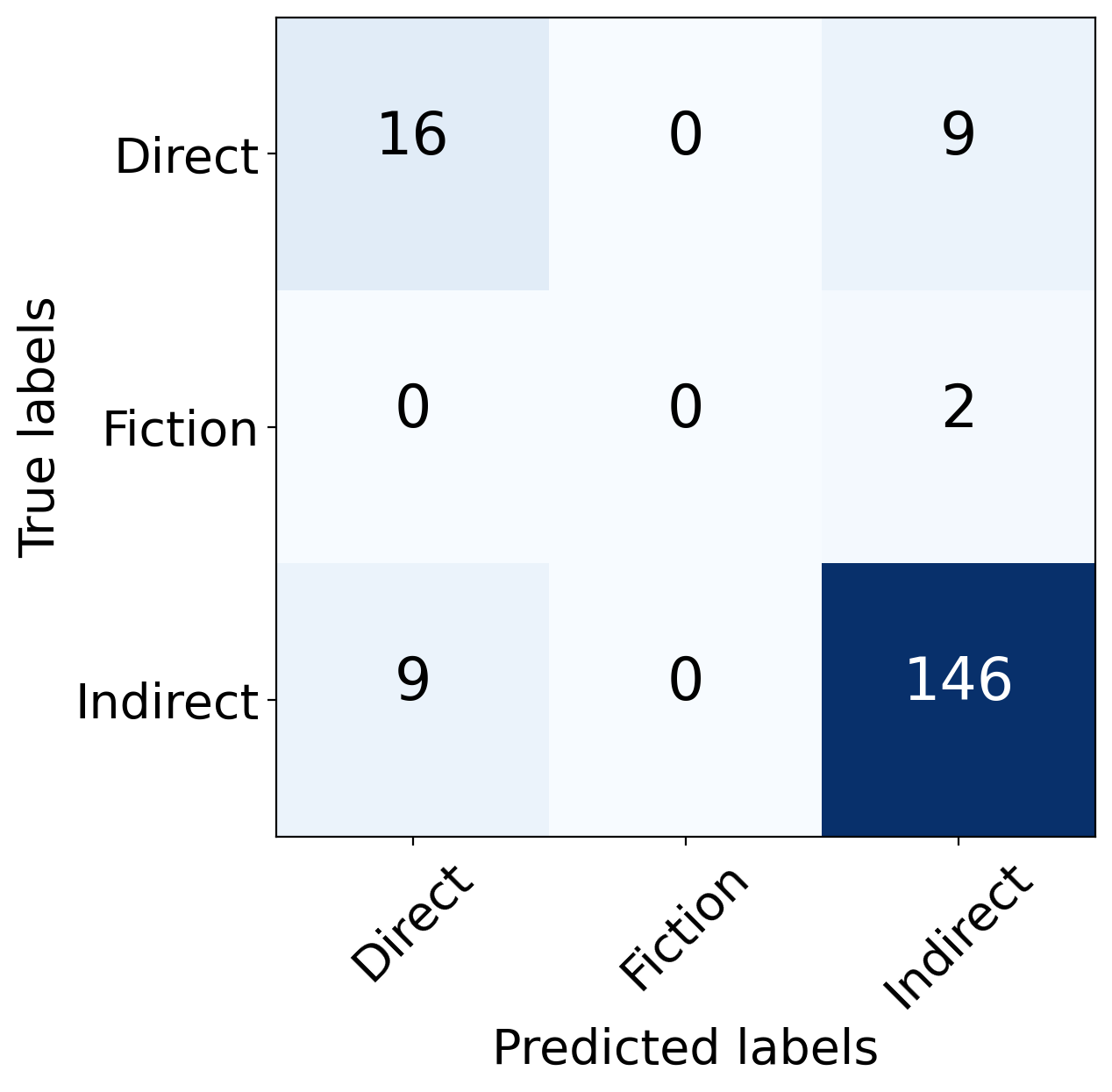}
        \caption{Subject}
        \label{fig:confusion-matrix-sog}
    \end{subfigure}
    \vspace{1cm}
    \begin{subfigure}{0.4\textwidth}
        \centering
        \includegraphics[width=\textwidth]{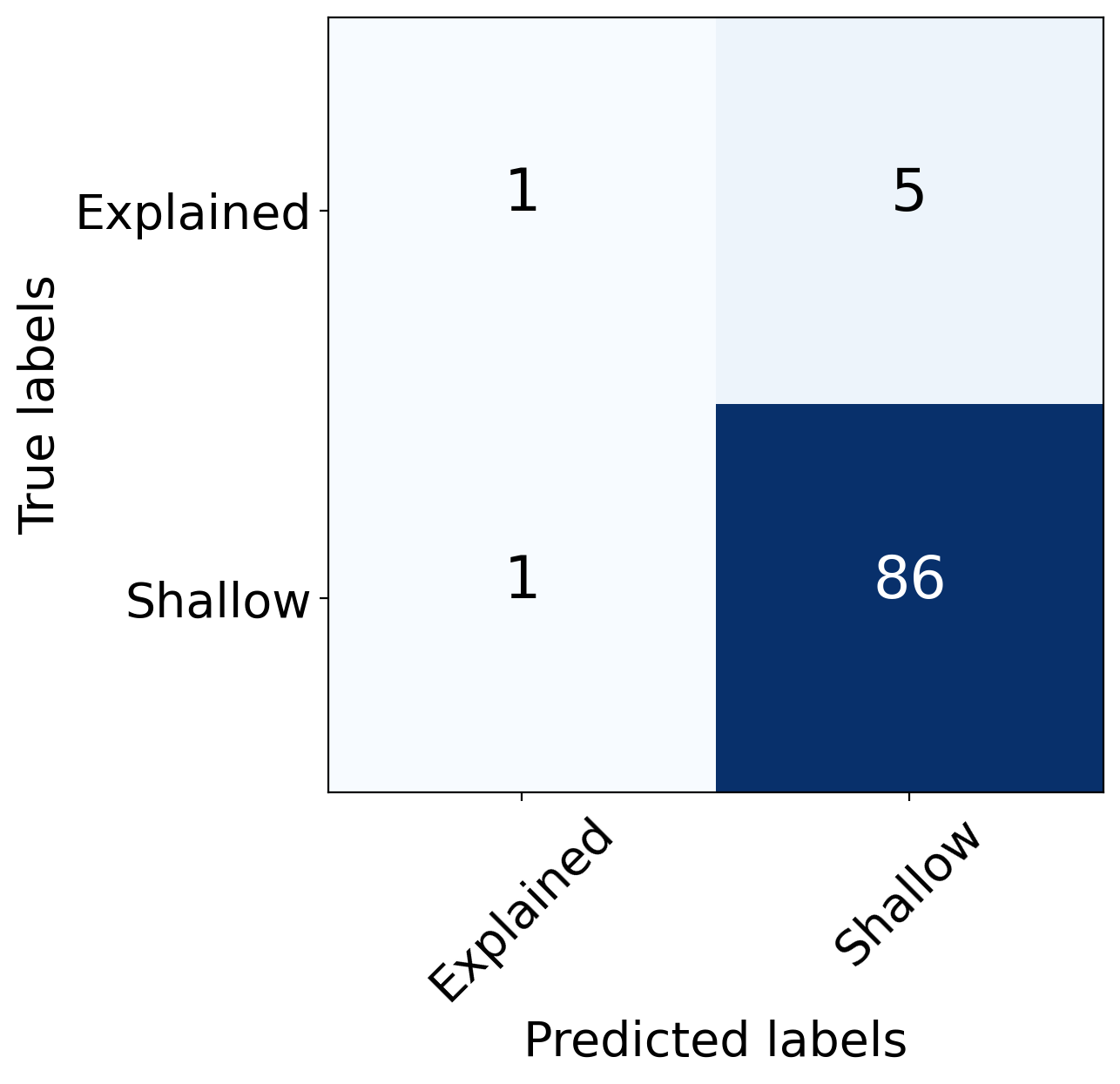}
        \caption{Argumentation}
        \label{fig:confusion-matrix-arg}
    \end{subfigure}
    \caption{Confusion Matrices between true labels and labels predicted by GPT-4 for Content Treatment Detection on the RAI-CMM-OPEN2 dataset}
    \label{fig:confusion-matrices}
\end{figure}

As defined in section\ref{contenttreatmentanalysi}, Content Treatment Analysis process defined by us in this work that assesses how content related to statements or topics is handled by the media or other sources. This involves categorizing content in terms of ``Orientation'' (reinforcing, confuting, or neutral), ``Subject'' (direct, indirect, or fiction), and `Argumentation'' (expained or shallow). The goal is to understand how information is presented and interpreted in a media context.

Content Treatment Detection refers to automatize this process with the support of large language models. At this purpose, we tested the performance of GPT-4 in recognizing different strata within the Content Treatment Detection task reporting both accuracies and balanced accuracies in Table \ref{tab:model-accuracy-RAI-CMM-OPEN2}.  ``All'' column indicates when all strata categories are recognized correctly, and it is not a simple average of the other individual columns. 

In addition Figure \ref{fig:confusion-matrices} shows the confusion matrices for each category serving as a visual representation for  understanding how well the model distinguishes between different classes.

The performance for the ``Subject'' and ``Argumentation'' exhibited notable effectiveness. However, it's important to note that in both cases, the minority classes were recognized less accurately, leading to a notable discrepancy between accuracy and balanced accuracy. Specifically, for the ``Subject'' stratum, GPT-4 always failed to recognize the ``Fiction'' class and this is the discrepancy cause, while for the ``Argumentation'' stratum the class ``Explained'' is predicted only 2 times out of 6.

All three strata labels were recognized as those annotated in only one-third of the cases. Further discussion on the discrepancy between the annotations and GPT-4 responses is provided in the following section.


\section{Large Language Models to evaluate text clarity}
\label{sec:textclaritycheck}

One of the most interesting applications of Large Language Models in this context could be the text clarity check. For instance, when considering the debunking of statements made by politicians, as done by fact-checking organizations like Pagella Politica, it is not always clear from the provided explanations what the verdict is. Take the following example: 

\texttt{Statement}: \textit{In Italy, there is a 12-13 percent of working poor, below the poverty line.}

\texttt{Text}: \textit{According to Eurostat, in 2020, workers in Italy living below the relative poverty threshold were 10.8 percent of the total. However, this percentage could be an underestimate.}

In this case, the verdict from Pagella Politica was ``True'', meaning the statement is correct. However, the provided explanation does not seem sufficient to support this verdict. The purpose of this example is not to criticize the work of Pagella Politica negatively (since their website provides both a brief explanation as shown above and a longer article), but rather to illustrate how Large Language Models could be used as a clarity check for content, similar to how Grammarly checks grammar and language. Querying Chat-GPT for previous example using prompt 1, we got the following response:

\textit{2 (not enough info) The text provides a percentage of 10.8\% of workers living below the relative poverty threshold in Italy in 2020 but does not provide specific information to confirm or refute the statement, which indicates a rate of 12-13\%. Additionally, the text suggests that the percentage could be an underestimate, but it does not offer additional data to establish with certainty whether this is the case or not.}

Given these reasons, the response is categorized as "Not Enough Info" because it acknowledges the presence of conflicting information and a lack of concrete evidence to fully support or refute the statement. Therefore, it seems to provide valid justifications for this classification.

In addition, during dataset annotation, especially when dealing with a limited number of annotators, as in our case, conducting a review and reassessment of the annotations can prove to be highly beneficial. For instance, in the case of Content Treatment Detection, a significant disparity was observed between human annotation and the output from GPT-4, particularly concerning the element of Orientation. This underscores the importance of considering AI-assisted review tools as a valuable resource to ensure the reliability of annotations.

\section{FAKE RADAR Game}
\label{sec:fakeradar}

In addition to our technical contributions, we actively participated in the development of the FAKE RADAR game, a valuable tool for raising awareness about fake news. It offers a variety of question types to challenge participants' ability to recognize and understand misinformation. Here's an itemized list of the different types of questions in the game:

\begin{itemize}
    \item \textbf{Textual Statements:} Some questions ask users to evaluate whether certain textual statements are true or false. For example, a question might present a statement like, ``In London, a man dressed as Vincent Van Gogh entered the headquarters of an environmental organization and threw minestrone soup at 'climate activists,' in response to an incident in London in October where environmentalists threw soup at Van Gogh's painting 'Sunflowers.''
    
    \item \textbf{Graph-Based Questions:} Questions present graphs and challenge players to deduce whether accurate conclusions can be drawn from the data presented in the graphs. For example, these graphs might be related to critical topics like climate change, and players must determine if they can infer specific information from the presented data. This type of question fosters competence in analyzing visual information and assessing complex data.
    
    \item \textbf{Multiple-Choice Questions:} Multiple-choice questions cover various topics related to disinformation. Users must select the correct or most appropriate answer from the options provided. This helps assess general knowledge of topics related to fake news and disinformation.
    
    \item \textbf{Images of People:} Images of people are displayed, some of which may be real, while others could be generated by artificial intelligence. Users must identify images of real people, aiding in the development of the ability to recognize manipulated images.
    
    \item \textbf{Film Clips with Deepfakes:} The game also features film clips in which deepfake technology may or may not have been used. For instance, in the documentary ``Welcome to Chechnya'' deepfakes were employed to protect the identities of LGBTQ+ community members providing testimony about the persecutions they have faced. This challenges players to distinguish between authentic and manipulated video content.

\end{itemize}
The complete list of questions is presented in these slides: \url{https://bitly.ws/X6W8}.

In a noteworthy addition, we introduced a version of the game at the Salone del Libro, where Chat-GPT also participated as a rival to the audience, responding to questions and competing in knowledge assessments. It's worth noting that Chat-GPT happened to win against the audience on some occasions, but it is essential to emphasize that this does not certify Chat-GPT as a source of absolute truth. In fact, during the event, concerns were raised about its potential use in creating fake news. 
On this last point, with respect to the verdicts issued by Pagella Politica on politicians' statements, Chat-GPT was tasked with generating alternative verdicts, and the results can be effortlessly incorporated into the game's question list. This facilitates the introduction of multiple-choice questions related to political fake news, with the correct verdict being determined by the journalists at Pagella Politica.

This game has been applied on various occasions beyond the Salone del Libro, such as Rai Porte Aperte and the Giffoni Film Festival, with the participants being middle and high school students. It would be interesting to consider expanding the game to a broader audience and analyzing how responses vary depending on the target demographic. Perhaps maintaining as future work an online instance of the game and regularly adding new questions could be a strategy. The quiz format is a versatile tool that can help individuals actively engage with important topics, even in informal settings like family gatherings, and we believe it's a valuable way to spread awareness about misinformation.

\section{Conclusion and future work}

The datasets and the code of this work can be downloaded from this GitHub repository even though it's being updated: \url{https://github.com/Loricanal/IDMO_AI_experiments.git}.

In summary, this project has yielded a range of significant outcomes and identified promising avenues for future exploration. These key achievements include the development of novel datasets for testing advanced technologies in fake news detection.

Furthermore, the project has made significant progress by introducing automatic models capable of predicting Pagella Politica's verdicts.

Notably, the project has achieved highly satisfactory results in the field of Textual Entailment Recognition, especially for the FEVER dataset. The creation of a \textit{DistilBERT} multilingual model has attained accuracy rates exceeding 90\%, surpassing the algorithm tested in the FEVER paper. 
With respect to Recognize Textual Entailment task, this work has also proposed a brief qualitative analysis of the annotations, highlighting some discrepancies among the datasets.

Finally, an exciting addition to the project has been the development of an engaging and informative game aimed at raising awareness about the widespread issue of fake news. This game has been showcased at various national events, contributing to the education and information of young audiences.

These accomplishments represent progress in the fields of fake news detection, media influence analysis, and automated content assessment. As we look to the future, our focus will be on building upon these achievements to enhance the project's effectiveness and extend its reach even further.
As we look forward to the future of this project, we have outlined several key areas for further development and enhancement. These future endeavors will not only contribute to the ongoing success of the project but also expand its scope and impact. The following points highlight the major areas of focus for future work:

\begin{enumerate}
    \item \textbf{Pagella Politica Tool to explore categorized verdicts:} Design a user-friendly tool that simplifies the process of merging multiple categories into one, allowing for comprehensive analyses of the accuracy of political parties and politicians. 
    
    \item \textbf{Langchain-Based Content Retrieval System:} Test a content retrieval system for fake news that relies on the Langchain technology. This innovative approach could revolutionize how we access and analyze information regarding misinformation. 
    
    \item \textbf{Fine-Tuning with GPT-3.5-turbo and GPT-4:} Explore the potential of fine-tuning the GPT-3.5-turbo and GPT-4 models for RTE and Textual Content Treatments tasks. 
    
    \item \textbf{FAKE RADAR game maintenance:} Ensure that the FAKE RADAR game remains current and engaging by continually enriching it with new questions, collecting users'answers.
\end{enumerate}

As our final point, it would be fascinating to explore the realm of continuous value scales, alongside the traditional discrete classes, for both verdict classification and textual entailment recognition. This approach would recognize the subtle shades of truth that exist beyond a binary framework.

\section*{Acknowledgments}
This work has been supported by the European Union's Connecting Europe Facility (CEF), under grant agreement nr. INEA/CEF/ICTA/20202394428 (IDMO).

\bibliographystyle{unsrt}  
\bibliography{references}  

\begin{thebibliography}{1}

\bibitem{FEVERPaper}
James Thorne, Andreas Vlachos, Christos Christodoulopoulos, and Arpit Mittal.
\newblock {FEVER}: a large-scale dataset for fact extraction and {VER}ification.
\newblock In {\em Proceedings of the 2018 Conference of the North {A}merican Chapter of the Association for Computational Linguistics: Human Language Technologies, Volume 1 (Long Papers)}, pages 809--819, New Orleans, Louisiana, June 2018. Association for Computational Linguistics.

\bibitem{WoodThomas}
Thomas Wood and Ethan Porter.
\newblock The elusive backfire effect: Mass attitudes’ steadfast factual adherence.
\newblock {\em Political Behavior}, 41, 03 2019.

\bibitem{santilli2023camoscio}
Andrea Santilli and Emanuele Rodolà.
\newblock Camoscio: an italian instruction-tuned llama, 2023.

\bibitem{hu2022lora}
Edward~J Hu, yelong shen, Phillip Wallis, Zeyuan Allen-Zhu, Yuanzhi Li, Shean Wang, Lu~Wang, and Weizhu Chen.
\newblock Lo{RA}: Low-rank adaptation of large language models.
\newblock In {\em International Conference on Learning Representations}, 2022.

\bibitem{DistilBERTPaper}
Victor Sanh, Lysandre Debut, Julien Chaumond, and Thomas Wolf.
\newblock Distilbert, a distilled version of {BERT:} smaller, faster, cheaper and lighter.
\newblock {\em CoRR}, abs/1910.01108, 2019.

\bibitem{bert}
Jacob Devlin, Ming{-}Wei Chang, Kenton Lee, and Kristina Toutanova.
\newblock {BERT:} pre-training of deep bidirectional transformers for language understanding.
\newblock In Jill Burstein, Christy Doran, and Thamar Solorio, editors, {\em Proceedings of the 2019 Conference of the North American Chapter of the Association for Computational Linguistics: Human Language Technologies, {NAACL-HLT} 2019, Minneapolis, MN, USA, June 2-7, 2019, Volume 1 (Long and Short Papers)}, pages 4171--4186. Association for Computational Linguistics, 2019.

\end{thebibliography}
\newpage
\begin{table}[!ht]
\centering
\caption{5 Random Sample Annotations from  FEVER dataset}
\vspace{0.33cm}
\label{samplesFEVER}
\begin{tabular}{p{6cm}p{6cm}l}
\hline
\textbf{Claim} & \textbf{Text} & \textbf{Label} \\
\hline
Stephen Colbert is the host of The Late Show on CBS. & Colbert has hosted The Late Show with Stephen Colbert, a late-night television talk show on CBS, since September 8, 2015. & Supported \\
Private Lives was created only by a Mexican soccer team. & Private Lives is a 1930 three-act costume comedy by Noël Coward. Sir Noël Peirce Coward (16 December 1899 – 26 March 1973) was an English playwright, composer, director, actor, and singer, known for his wit, flamboyance, and what Time magazine called ``a sense of personal style, a combination of cheek and chic, pose and poise.'' & Refuted \\
Sebastian Stan acted in Political Animals. & \cellcolor{yellow!50} His role in Political Animals earned him a nomination for the Critics' Choice Television Award for Best Supporting Actor in a Movie/Miniseries. & Supported \\
Anne Rice lives in San Francisco. & \cellcolor{yellow!25} Born in New Orleans, Rice spent most of her early years there before moving to Texas and later to San Francisco. & Supported \\
The Prowler made his first appearance in The Amazing Spider-Man \# 68. & Created by writer-editor Stan Lee, John Buscema, and Jim Mooney, Prowler made his first appearance in The Amazing Spider-Man \#78. & Refuted \\
\hline
\end{tabular}
\end{table}


\begin{table}[!ht]
\centering
\caption{5 Random Sample Annotations from  Ministery of Health dataset}
\vspace{0.33cm}
\label{samplesFNM}
\begin{tabular}{p{4cm}p{8cm}l}
\hline
\textbf{Claim} & \textbf{Text} & \textbf{Label} \\
\hline
The soles of shoes definitely carry the virus into the home and can transmit the infection & The probability that SARS-CoV-2 can spread through shoes and infect individuals is very low. As a precaution, especially in homes where infants and children crawl or play on the floor, it is possible to leave shoes at the entrance to the house. This way, contact with dirt that could be carried on the soles of shoes will be avoided. & Refuted \\
SARS-CoV-2 infection can be contracted from your pet & There is no scientific evidence that domestic animals, such as dogs and cats, can transmit the new coronavirus to humans. As a general rule of hygiene, it is recommended to wash hands thoroughly with soap after contact with animals, a common practice to protect against other microorganisms that can be transmitted from animals to humans. & Refuted \\
If I eat more proteins, I produce more antibodies and boost my immune defenses & No, there is currently no scientific evidence that increasing protein intake through diet provides benefits to the immune system. Caution should always be exercised before starting a high-protein diet, as this type of diet can have health consequences (such as the development of kidney diseases), and excessive consumption of meat and proteins can increase the risk of developing certain forms of cancer. DIY is always to be avoided. A varied and balanced diet is the basis for a healthy lifestyle. & Refuted \\
COVID-19 spreads through breath and saliva droplets, so there is no need to disinfect objects and surfaces & SARS-CoV-2 primarily spreads through respiratory droplets, including aerosols from an infected person who sneezes, coughs, speaks, sings, or breathes near others. Droplets, including aerosols, can be inhaled or deposited in the nose, mouth, or eyes. Therefore, ensuring ventilation in enclosed spaces is extremely important. Less commonly, infection may be due to contact with surfaces contaminated by droplets. This means that if we touch objects or surfaces with respiratory secretions from infected individuals and then bring them to our mouth, nose, or eyes, we can become infected indirectly. In addition to frequent hand washing, regular cleaning and disinfection of surfaces, especially frequently touched surfaces (such as handles, handrails, remote controls, light switches, etc.), are recommended as a prevention and control measure against the spread of the virus. & Refuted \\
A beard exposes to a higher risk of getting infected with the new coronavirus & The beard itself does not expose to a higher risk of contracting the new coronavirus compared to those without it. Of course, its hygiene should be taken care of, and when wearing a mask, it should be adjusted to adhere well to the face. There is no evidence that shaving the beard helps avoid SARS-CoV-2 infection. The recent news on this topic stems from the hasty interpretation of a 2017 infographic from the Centers for Disease Control and Prevention, dedicated to workplace safety and the correct use of masks in general. The CDC's recommendations refer to the correct use of medical protective equipment, comparing 36 types of shaving, 18 of which are not recommended because they could render the mask ineffective. & Refuted \\
\hline
\end{tabular}
\end{table}

\begin{table}[!ht]
\centering
\caption{5 Random Sample Annotations from  PagellaPolitica1 dataset}
\vspace{0.33cm}
\label{samplesPagellaPolitica1}
\begin{tabular}{p{6cm}p{6cm}l}
\hline
\textbf{Claim} & \textbf{Text} & \textbf{Label} \\
\hline
\cellcolor{yellow!50}"We don't spend money to buy weapons that we send to the Ukrainians." & So far, Italy has sent six shipments of weapons to Ukraine: the lists are not public, so it is not possible to know their value. However, we know that they are weapons already available to the Italian armed forces, so they are not new. Minister Crosetto has already said that it will be necessary to replenish the stocks of weapons given the aid sent to Ukraine, and this will have a cost. Additionally, Italy contributes to the financing of the "European Peace Facility," a EU fund that has allocated 3.6 billion euros for military assistance to Ukraine so far. & Refuted \\
\cellcolor{yellow!50} "I [didn't vote] for this electoral law, and I didn't vote for the previous one either." & The current electoral law, the Rosatellum, was approved in 2017 when Letta did not hold any parliamentary office. In 2015, Letta did not participate in three votes of confidence related to the Italicum and resigned shortly after its approval. & Supported \\
\cellcolor{orange!50} "The water network wastes about 40 percent of our water." & According to Istat, in 2018, 42 liters of water were wasted for every 100 liters introduced into the national water network. Most of the losses are due to the poor condition of the infrastructure. & Supported \\
"The League's proposal [for a 15 percent flat tax] has 18 brackets." & The League's bill for a 15 percent flat tax includes a transitional phase with 18 brackets for taxing the incomes of families and individuals. & Supported \\
"The new role of Di Maio was had by Tony Blair." & Di Maio's new role is unique and concerns energy and security issues with countries such as Saudi Arabia, Qatar, and the United Arab Emirates. Between 2007 and 2015, Blair was a special envoy for the Middle East Quartet and dealt with the Israel-Palestine conflict. & Refuted \\
\hline
\end{tabular}
\end{table}

\end{document}